\documentclass[letterpaper, 10 pt, conference]{ieeeconf}  

\IEEEoverridecommandlockouts                              

\usepackage{graphics} 
\usepackage{graphicx}

\usepackage{epsfig} 
\usepackage{amsmath} 
\usepackage{amssymb}  
 \usepackage{algorithm}
\usepackage{algpseudocode}
\algnewcommand\AAND{\textbf{ and }}
\algnewcommand\Or{\textbf{ or }}
\usepackage{color}
\usepackage{citesort}
\usepackage{url}

\DeclareMathAlphabet{\pazocal}{OMS}{zplm}{m}{n}

\newcommand{\notab}{\noindent}

\newcommand{\As}{\pazocal{A}}

\newcommand{\Rs}{\pazocal{R}}

\newcommand{\Ss}{\pazocal{S}}

\usepackage{array}
\newcolumntype{C}[1]{>{\centering\arraybackslash}p{#1}}
\newcolumntype{M}[1]{>{\raggedright\arraybackslash}p{#1}}

\usepackage{array} 
\newcolumntype{L}[1]{>{\raggedright\let\newline\\\arraybackslash\hspace{0pt}}m{#1}}	
\newcolumntype{S}[1]{>{\centering\let\newline\\\arraybackslash\hspace{0pt}}m{#1}}
\newcolumntype{R}[1]{>{\raggedleft\let\newline\\\arraybackslash\hspace{0pt}}m{#1}}

\algnewcommand\pushup{\vspace{-1ex}}
\algnewcommand\pushuphalf{\vspace{-0.5ex}}

\makeatletter
\renewcommand*{\@opargbegintheorem}[3]{\trivlist
  \item[\hskip \labelsep{\itshape #1\ #2}] \textit{(#3)}\ }
\makeatother
\title{\LARGE \bf
Model Predictive Control for Micro Aerial Vehicles: A Survey
}

\author{Huan Nguyen$^{1}$, Mina Kamel$^{2}$, Kostas Alexis$^{1}$, and Roland Siegwart$^{3}$
\thanks{$^{1}$The authors are with the Autonomous Robots Lab, Norwegian University of Science and Technology (NTNU), 7491, Trondheim, Norway
        {\tt\small huan.nguyen(konstantinos.alexis)@ntnu.no}}
\thanks{$^{2}$M. Kamel is with Voliro AG, Weinbergstrasse 35, 8092, Zurich, Switzerland
        {\tt\small mina.kamel@voliro.com}}%
\thanks{$^{3}$R. Siegwart is with the Autonomous Systems Lab, ETH Zurich, Leonhardstrasse 21, 8092, Zurich, Switzerland
        {\tt\small rsiegwart@ethz.ch}}%
}

\begin{document}

\maketitle
\thispagestyle{empty}
\pagestyle{empty}

\begin{abstract}

This paper presents a review of the design and application of model predictive control strategies for Micro Aerial Vehicles and specifically multirotor configurations such as quadrotors. The diverse set of works in the domain is organized based on the control law being optimized over linear or nonlinear dynamics, the integration of state and input constraints, possible fault-tolerant design, if reinforcement learning methods have been utilized and if the controller refers to free-flight or other tasks such as physical interaction or load transportation. A selected set of comparison results are also presented and serve to provide insight for the selection between linear and nonlinear schemes, the tuning of the prediction horizon, the importance of disturbance observer-based offset-free tracking and the intrinsic robustness of such methods to parameter uncertainty. Furthermore, an overview of recent research trends on the combined application of modern deep reinforcement learning techniques and model predictive control for multirotor vehicles is presented. Finally, this review concludes with explicit discussion regarding selected open-source software packages that deliver off-the-shelf model predictive control functionality applicable to a wide variety of Micro Aerial Vehicle configurations.

\end{abstract}

\section{INTRODUCTION}\label{sec:into}
Micro Aerial Vehicles (MAVs) and especially systems of the multirotor class, such as quadrotors and hexacopters, correspond to a widely adopted type of aerial robot. Such systems are nowadays extensively used for autonomous inspection~\cite{SIP_AURO_2015}, surveillance~\cite{grocholsky2006cooperative} and other remote sensing applications, alongside tasks relating to physical interaction~\cite{garimella2015towards}, delivery~\cite{agha2014health} and more. Their success is attributed to a variety of factors including their simplicity, low-cost, reliability, and agile dynamics. Naturally, a key component relates to the accuracy and robustness of the controller onboard such systems which alongside the state estimation process are the two most fundamental algorithms necessary to facilitate autonomous navigation. 

%
\begin{figure}[h!]
\centering
    \includegraphics[width=0.99\columnwidth]{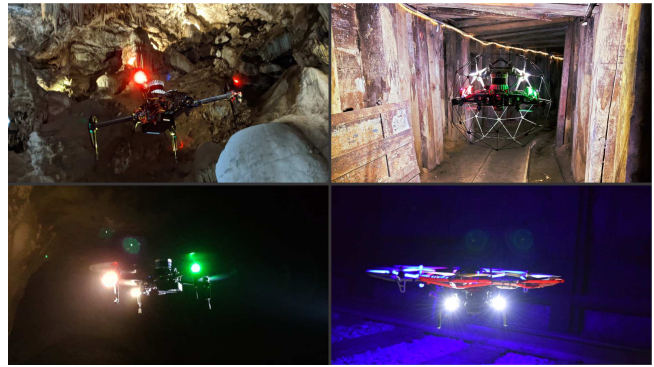}
\caption{Indicative robots from previous work of the authors that have relied on model predictive control for their position control.}\label{fig:motivation}
\vspace{-4ex} 
\end{figure}
%

In response to this fact, a wide variety of control strategies have been proposed for the problem of MAV flight control including both model-free and model-based methods. In the latter, both linear and nonlinear methods have been considered, alongside methods exploiting piecewise system models, techniques tailored to robots undergoing physical interaction, load transportation, and deep neural networks-based reinforcement learning approaches. Among the multiple approaches, model predictive control has seen wide utilization and has presented outstanding results in terms of trajectory tracking accuracy and robust performance. Figure~\ref{fig:motivation} presents examples of MAVs relying on model predictive control.

Model Predictive Control (MPC)~\cite{camacho2013model,allgower2012nonlinear,mayne2000constrained,grune2017nonlinear,garcia1989model,rawlings2000tutorial,bemporad1999robust} offers a collection of properties of significant importance for MAVs. Being a model-based method, it can exploit knowledge of the dynamics model of the system. Based on the extensive progress in the domain, MPC methods are now feasible both for linear and nonlinear systems, alongside hybrid model formulations. By optimizing over a horizon, MPC can simultaneously optimize towards optimal tracking of the reference trajectory and satisfy input and state constraints, while retaining robust performance. Furthermore, state constraints may not be limited to box constraint formulations but also model $3\textrm{D}$ obstacles as regions of the navigation space that must be avoided. Additionally, MPC by nature relates to approximate dynamic programming and is very relevant to modern research in reinforcement learning, a fact reflected in a multitude of new works of the community. Moreover, the power of MPC has enabled it to solve complex problems in MAV autonomy such as the recent perception-aware model predictive navigation method in~\cite{falanga2018pampc}. 

In this paper we provide a survey with respect to the methods proposed for trajectory tracking control of MAVs of quadrotor, hexarotor and other multirotor configurations. We cover the domains of Linear Model Predictive Control (LMPC) and Nonlinear MPC (NMPC), as well as MPC for aerial manipulation and load transportation, fault-tolerant control, alongside the interconnection between MPC and neural networks-based reinforcement learning approaches. We present selective comparison results which serve to provide design guidelines and further categorize a set of open-source code packages that provide off-the-shelf functionality for deploying MPC onboard micro aerial vehices. 

The rest of this paper is organized as follows. A model of the multirotor dynamics is overviewed in Section~\ref{sec:modeling}. The survey presentation of MPC for MAVs is detailed in Section~\ref{sec:survey} with subsections on linear and nonlinear methods, strategies for fault-tolerance, load transportation, physical interaction and works involving deep reinforcement learning. Finally, Section~\ref{sec:opensource} outlines a selected set of open-source packages, while conclusions are drawn in Section~\ref{sec:conclusion}. 

\section{MODELING OF MICRO AERIAL VEHICLES}\label{sec:modeling}
\begin{figure}[h!]
\centering
    \includegraphics[width=0.8\columnwidth]{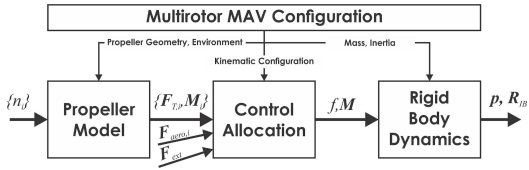}
\caption{Basic model components of MAV dynamics.}\label{fig:modelcomponents}
\vspace{-2ex} 
\end{figure}

A set of contributions have provided extensive means to model multirotor MAVs at selective levels of fidelity. As visually depicted in Figure~\ref{fig:modelcomponents}, one may account to a different extent for complex aerodynamic parameters, non-diagonal inertia terms and other effects that have been detailed extensively in pioneering studies~\cite{hoffmann2007quadrotor}. This modular approach allows us to simplify, without loss of generality, the subsequent discussion by considering the hexarotor vehicle as a particular instance of a multirotor system, while researchers that build upon this presentation may decide independently of components such as the propeller model. A hexarotor is typically a platform consisting of six identical rotors and propellers symmetrically configured. This propulsion system generates the thrust and torque normal to the plane of the vehicle, as required to facilitate stable control. 

%
%

\begin{figure}[h!]
\centering
    \includegraphics[width=0.85\columnwidth]{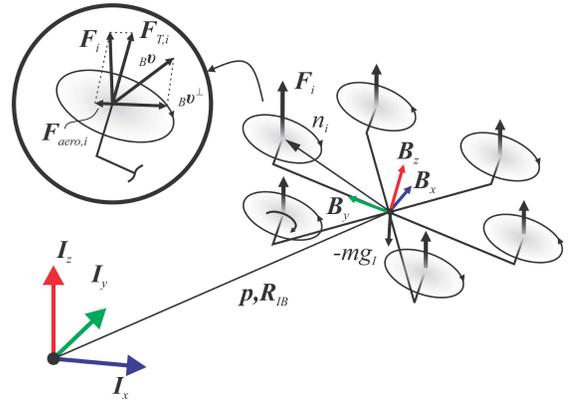}
\caption{Hexarotor model and utilized coordinate frames.}\label{fig:coordframes}
\vspace{-2ex} 
\end{figure}

For the modeling derivations below we choose an inertial reference frame $\mathbb{I}$ with unit vectors $\{\vec{\mathbf{I}}_x,\vec{\mathbf{I}}_y,\vec{\mathbf{I}}_z\}$ and a body fixed frame $\mathbb{B}$ with unit vectors $\{\vec{\mathbf{B}}_x,\vec{\mathbf{B}}_y,\vec{\mathbf{B}}_z\}$. The origin of $\mathbb{B}$ is located at the Center of Mass (CoM) of the hexarotor and is presented in Figure~\ref{fig:coordframes}. For the rest of this process, let us denote $m$ as the total mass, $\mathbf{J}\in{\mathbb{R}}^{3 \times 3}$ the inertia matrix with respect to $\mathbb{B}$, $\mathbf{R}_{IB}\in SO(3)$ the rotation matrix representing the vehicle orientation, $\boldsymbol{\omega}\in {\mathbb{R}}^3$ the angular velocity expressed in $\mathbb{B}$, $\mathbf{p}\in {\mathbb{R}}^3$ the position of the vehicle's CoM in expressed in $\mathbb{I}$, and $\boldsymbol{\upsilon}\in{\mathbb{R}}^3$ the velocity of the CoM expressed in $\mathbb{I}$.

The dominant forces acting on the vehicle are generated from the propellers. Under a set of common and well-proven assumptions, each propeller is considered to generate thrust proportional to the square of the propeller rotation speed and angular moment due to the drag force. For each propeller $i$, the generated thrust and moment take the form:

\small
\begin{eqnarray}\label{eq:prop}
\mathbf{F}_{T,i} &=& k_n n_i^2 \mathbf{e}_z \\ \nonumber
\mathbf{M}_{i} &=& (-1)^{i-1}k_m \mathbf{F}_{T,i}
\end{eqnarray}
\normalsize
where $n_i$ is the rotor speed of the propeller, $k_n,k_m>0$ are constants, and $\mathbf{e}_z$ is a unit vector in the $z$ direction. 

%
%

This level of modeling fidelity for the forces applied on a multirotor is the one most commonly found. However, if we aim to consider dynamic maneuvers, then two additional phenomena come into play. These effects are the blade flapping and induced drag and introduce additional forces in the $x\textrm{-}y$ rotor plane and thus add more damping to the MAV~\cite{mahony2012multirotor}. Combining these effects into one lumped drag coefficient~\cite{omari2013nonlinear}, we derive the following aerodynamic force for propeller $i$: 

\small
\begin{eqnarray}
\mathbf{F}_{aero,i} &=& f_{T,i} \mathbf{K}_{drag} \mathbf{R}_{IB}^T \mathbf{v}
\end{eqnarray}
\normalsize
where $\mathbf{K}_{drag}=\textrm{\textbf{diag}}(k_D,k_D,0),~k_D>0$, and $f_{T,i}$ is the $z$-component of the $i$-th thrust force. Then the motion dynamics take the form: 

\scriptsize
\begin{eqnarray}
\dot{\mathbf{p}} &=& \boldsymbol{\upsilon} \label{eq:position}\\ 
\dot{\boldsymbol{\upsilon}} &=& \frac{1}{m}\left ( \mathbf{R}_{IB}\sum_{i=0}^{N_r}\mathbf{F}_{T,i} \!-\! \mathbf{R}_{IB}\sum_{i=0}^{N_r}\mathbf{F}_{aero,i}\!+\!\mathbf{F}_{ext} \right ) \!+\! \begin{bmatrix}
0\\ 
0\\ 
-g
\end{bmatrix} \label{eq:velocity}\\ 
\dot{\mathbf{R}}_{IB} &=& \mathbf{R}_{IB}\left \lfloor \boldsymbol{\omega} \times \right \rfloor \label{eq:rotmat}\\ 
\mathbf{J}\dot{\boldsymbol{\omega}} &=& -\boldsymbol{\omega}\times \mathbf{J} + {\boldsymbol{\As}} \begin{bmatrix}
n_1^2\\ 
\vdots\\ 
n_{N_r}^2
\end{bmatrix} \label{eq:attirates}
\end{eqnarray}
\normalsize
where $\mathbf{F}_{ext}$ represents any external forces acting on the vehicle, and $\boldsymbol{\As}$ is the control allocation matrix and $N_r$ the number of propellers. The works in~\cite{kamel2015fast,lee2010geometric} present the control allocation matrix derivations for the case of symmetric hexarotor and quadrotors respectively.

\notab \textit{Attitude Subsystem:} It is noted that commonly in application, the attitude dynamics of a multirotor platform are controlled with a fast embedded system running a rather simple to calculate feedback loop often only involving fixed-gains. Therefore, MPC is often deployed as a cascale position controller commanding the closed loop attitude dynamics which now should be identified. For that goal, the inner-loop attitude model can be represented as a first-order model due to the efficiency of onboard control and despite its otherwise second-order nature~\cite{kamel2017linear}. The closed-loop attitude dynamics to then be identified take the form: 

\small
\begin{eqnarray}\label{eq:attid}
\dot\phi &=& \frac{1}{\tau_\phi}(k_\phi \phi_{ref}-\phi) \\ 
\dot\theta &=& \frac{1}{\tau_\theta}(k_\theta \theta_{ref} - \theta) \nonumber\\ 
\dot\psi &=& \dot\psi_{ref} \nonumber
\end{eqnarray}
\normalsize
where $k_\phi,k_\theta$ and $\tau_\phi,\tau_\theta$ are the dc-gains and time constants of the roll and pitch closed-loop dynamics respectively, while $\phi_{ref},\theta_{ref}$ represent the reference roll and pitch angles, and $\dot\psi_{ref}$ is the commanded yaw rate.

\section{MODEL PREDICTIVE CONTROL FOR MAVs}\label{sec:survey}
In this section we overview some of the successful methods and strategies of applying model predictive control for MAVs. In particular, linear and nonlinear schemes are presented, methods for physical interaction and load transportation, alongside techniques combining traditional MPC and neural networks-based reinforcement learning. 

\subsection{Linear Model Predictive Control}\label{sec:lmpc}

The basic case of application of MPC for quadrotor control relates to linear methods. Furthermore, in the most widely adopted case, Linear Model Predictive Control (LMPC) is deployed to handle the position dynamics of a MAV assuming that an attitude controller is already deployed and an associated closed-loop attitude dynamics model has been identified as described in Eq.~(\ref{eq:attid}). Given this model we can proceed to linearize the remaining system dynamics around hover. We define the following state vector and control input:

\small
\begin{eqnarray}
\mathbf{x} = [\mathbf{p}^T~\boldsymbol{\upsilon}^T~ _{\mathbb{I}} \phi~ _{\mathbb{I}} \theta ]^T \\ 
\mathbf{u} = [ _{\mathbb{I}} \phi_{ref}~ _{\mathbb{I}} \theta_{ref}~ T_{ref}]^T
\end{eqnarray}
\normalsize
where $T_{ref}$ is the commanded reference thrust, $_{\mathbb{I}} \phi, _{\mathbb{I}} \theta$
are the roll and pitch angles expressed in the inertial frame. The following relation with the robot roll and pitch angles holds:

\small
\begin{eqnarray}
\begin{bmatrix}
\phi \\ 
\theta 
\end{bmatrix} = \begin{bmatrix}
\cos \psi & \sin \psi \\ 
-\sin \psi & \cos \psi
\end{bmatrix} \begin{bmatrix}
_{\mathbb{I}} \phi\\ 
_{\mathbb{I}} \theta
\end{bmatrix}
\end{eqnarray}
\normalsize
Finally, after linearization and discretization the following state-space form holds in which the effect of external forces $\mathbf{F}_{ext,k}$ and the disturbance matrix $\mathbf{B}_d$ are also considered:

\small
\begin{eqnarray}
\mathbf{x}_{k+1} = \mathbf{A}\mathbf{x}_k + \mathbf{B}\mathbf{u}_k + \mathbf{B}_d\mathbf{F}_{ext,k}
\end{eqnarray}
\normalsize 

Provided the above, the LMPC strategy repeatedly solves the following Optimal Control Problem (OCP) assuming that input constraints apply but no state constraints are considered:

\small
\begin{align}
    \min _{\mathbf{U}}\sum_{k=0}^{N-1}\left ( \left \| \mathbf{x}_k - \mathbf{x}_{ref,k} \right \|^2_{\mathbf{Q}_x} + \left \| \mathbf{u}_k - \mathbf{u}_{ref,k} \right \|^2_{\mathbf{R}_u} \right ) \\ \nonumber + \left \| \mathbf{x}_N - \mathbf{x}_{ref,N} \right \|^2_\mathbf{P}
\end{align}
\begin{eqnarray}
    \textrm{s.t.} && \mathbf{x}_{k+1} = \mathbf{A}\mathbf{x}_k + \mathbf{B}\mathbf{u}_k + \mathbf{B}_d\mathbf{F}_{ext,k}\\ \nonumber
    && \mathbf{F}_{ext,k+1} = \mathbf{F}_{ext,k},~k=0,...,N-1\\ \nonumber
    && \mathbf{u}_k \in \mathbb{U}\\ \nonumber
    && \mathbf{x}_0 = \mathbf{x}(t_0),~\mathbf{F}_{ext,0}= \mathbf{F}_{ext}(t_0)
\end{eqnarray}
\normalsize
where $\mathbf{Q}_x\succeq 0, \mathbf{R}_u\succeq 0$ are the state and input penalty matrices, while $\mathbf{P} \succeq 0$ is the terminal state error penalty. Furthermore $\mathbf{x}_{ref,k},\mathbf{u}_{ref,k}$ are the target state and target control input $\mathbf{u}_{ref,k} = [_{\mathbb{I}}\phi_{ref,k}, ~_{\mathbb{I}}\theta_{ref,k},~T_{ref,k}]$ respectively at time $k$. The input constraints take the following form: 

\small
\begin{eqnarray}
\mathbb{U} = \begin{Bmatrix}
\mathbf{u} \in {\mathbb{R}^3} | \begin{bmatrix}
\phi_{\min}\\ 
\theta_{\min}\\ 
T_{ref,\min}
\end{bmatrix} \le \mathbf{u} \le \begin{bmatrix}
\phi_{\max}\\ 
\theta_{\max}\\ 
T_{ref,\max}
\end{bmatrix}
\end{Bmatrix}
\end{eqnarray}
\normalsize
Provided the derivation of the control law per iteration, the method then applies the first control input $\mathbf{u}_0$ and the whole process is repeated in a receding horizon fashion. Lastly, it is noted that the derived thrust reference vector is nonlinearly scaled to account for the projection of thrust when the system roll and pitch are nonzero: 

\small
\begin{eqnarray}
\tilde{T}_{ref} = \frac{T_{ref} + g}{\cos\phi \cos\theta}
\end{eqnarray}
\normalsize

\notab \textit{Disturbance Observer:} A disturbance observer can be incorporated to the above design for offset-free tracking. This is achieved by augmenting the system model with the disturbances vector. Considering the need to track the system output $\mathbf{y}_k = \mathbf{C}\mathbf{x}_k$ and achieve offset-free tracking, a simple observer to estimate such a disturbance takes the form:

\tiny
\begin{eqnarray}
\begin{bmatrix}
\hat{\mathbf{x}}_{k+1}\\ 
\hat{\mathbf{F}}_{ext,k+1}
\end{bmatrix} = \begin{bmatrix}
\mathbf{A} & \mathbf{B}_d\\ 
\mathbf{0} & \mathbf{I} 
\end{bmatrix}\begin{bmatrix}
\hat{\mathbf{x}}_{k}\\ 
\hat{\mathbf{F}}_{ext,k}
\end{bmatrix} + 
\begin{bmatrix}
\mathbf{B}\\ 
\mathbf{0}
\end{bmatrix}\mathbf{u}_k + 
\begin{bmatrix}
\mathbf{L}_x\\ 
\mathbf{L}_{F_{ext}}
\end{bmatrix}(\mathbf{C}\hat{\mathbf{x}}_k - \mathbf{y}_{m,k})
\end{eqnarray}
\normalsize
where $\hat{\mathbf{x}}_k, \hat{\mathbf{F}}_{ext,k}, \mathbf{y}_{m,k}$ are the estimated state, external disturbances and measured output at time $k$, respectively, while $\mathbf{L}_x,\mathbf{L}_{F_{ext}}$ are the associated observer gains. Assuming a stable observer, we can compute the steady-state MPC state $\mathbf{x}_{ref,k}$ and control input $\mathbf{u}_{ref,k}$ at time $k$ by solving:

\scriptsize
\begin{eqnarray}
\begin{bmatrix}
\mathbf{A}-\mathbf{I} & \mathbf{B} \\ 
\mathbf{C} & \mathbf{0}
\end{bmatrix}\begin{bmatrix}
\mathbf{x}_{ref,k}\\ 
\mathbf{u}_{ref,k}
\end{bmatrix} = \begin{bmatrix}
-\mathbf{B}_d \hat{\mathbf{F}}_{ext,k}\\ 
\mathbf{r}_k
\end{bmatrix}
\end{eqnarray}
\normalsize
where $\mathbf{r}_k$ the output vector reference at time $k$.

\notab \textit{Literature Review:} The abovementioned derivation corresponds to the most straightforward application of linear MPC for the position control of MAVs. At the same time the research community has explored a much more rich set of methods. Early in the timeline of this research, the authors in~\cite{alexis2010design} proposed the application of such a receding horizon scheme for the attitude control of a quadrotor vehicle and further accounted for state constraints. As the calculation of MPC subject to input and state constraints can be expensive - especially in comparison to the fast attitude dynamics - multiparametric approaches have been investigated for the explicit derivation of the control law~\cite{MPT3}. At a similar period, the authors in~\cite{raffo2010integral} proposed LMPC methods with integral terms. Aiming to account for the change in the system dynamics when the operating point departs significantly from the hovering point - but still not employing nonlinear methods - the works in~\cite{alexis2011switching,alexis2012model} present a PieceWise Affine (PWA) modeling approach and associated predictive control policy for the full control of a quadrotor MAV. Furthermore, the work in~\cite{alexis2016robust} investigated the design of robust MPC methods and presented extensive disturbance rejection capabilities including the ability to handle slung load disturbances. Currently, LMPC methods have presented significant success and have managed to be utilized reasonably extensively at least in multirotors in research labs as also visible in the discussion for open source packages in Section~\ref{sec:opensource}. Connecting the domain of linear and nonlinear MPC approaches, the work in~\cite{greeff2018flatness} offers a flatness-based approach which exploits feedback linearization and provides agile flight capabiltiies across the flight envelope but with the often reduced computational cost of linear methods. 

\notab \textit{Reachability Analysis:} When safety-critical applications are considered, guaranteed control performance is necessary. Generally, for a dynamic system, the reachable set $\Rs$ for a time $t$, inputs $u$, disturbances $w$ and a set of initial states $\Ss$ is the set of end states of trajectories starting in $\Ss$ after time $t$~\cite{schurmann2018reachset}. Despite the importance of reachable set analysis for MPC controllers, the literature in MPC application for MAVs mostly lacks such considerations. Few directly or indirectly relevant exceptions have examined the problem either directly from a MPC standpoint or with regards to learning-based methods~\cite{gillula2011applications,aswani2012extensions}, yet it is believed that the domain deserves further attention.

\subsection{Nonlinear Model Predictive Control}\label{sec:nmpc}
Linear control methods are appealing due to their simplicity and often reduced computational needs. Long experience in the community has indicated that when a multirotor MAV is largely operating around hovering/small-angles then LMPC methods provide high performance and robustness. However, nonlinear control has to be utilized if the complete flight envelope of the system is to be exploited. 

Towards that goal we derive a baseline formulation for Nonlinear Model Predictive Control. We consider the following state and control vectors: 

\small
\begin{eqnarray}
\mathbf{x} = [\mathbf{p}^T~\boldsymbol{\upsilon}^T~ _{\mathbb{I}} \phi~ _{\mathbb{I}} \theta ~_{\mathbb{I}} \psi ]^T \\ 
\mathbf{u} = [ _{\mathbb{I}} \phi_{ref}~ _{\mathbb{I}} \theta_{ref}~ T_{ref}]^T
\end{eqnarray}
\normalsize
This in turn allows us to formulate the nonlinear OCP: 

\small
\begin{eqnarray}\label{eq:nmpcocp}
\min_{\mathbf{U}}\int _{t=0}^T \left \| \mathbf{x}(t) -\mathbf{x}_{ref}(t) \right \|^2_{\mathbf{Q}_x} + \left \| \mathbf{u}(t)-\mathbf{u}_{ref}(t) \right \|^2_{\mathbf{R}_u}dt \\ \nonumber
+ \left \| \mathbf{x}(T)-\mathbf{x}_{ref}(T) \right \|^2_{\mathbf{P}}
\end{eqnarray}
\begin{eqnarray}
    \textrm{s.t.} && \dot{\mathbf{x}}=\mathbf{f}(\mathbf{x},\mathbf{u}) \\ \nonumber
&& \mathbf{u}(t) \in \mathbb{U} \\ \nonumber
&& \mathbf{x}(0) = \mathbf{x}(t_0)
\end{eqnarray}
\normalsize
where $\mathbf{f}$ is composed by Eqs.~(\ref{eq:position})~(\ref{eq:velocity})~(\ref{eq:attid}). The controller is implemented in a receding horizon fashion, where this optimization needs to be solved in real-time. As typically this corresponds to a computationally expensive task, especially for the fast dynamics of MAVs and the often limited onboard computational capabilities, direct methods~\cite{kamel2017linear} have gained significant attention due to their reduced processing needs. Multiple shooting techniques in particular have been used to solve Eq.~(\ref{eq:nmpcocp})~\cite{kamel2017linear} with the system dynamics and constraints being disccretized over a coarse discrete time grid $t_0,...,t_N$ within the interval $[t_k,t_{k+1}]$ and for each interval solving a Boundary Value Problem where additionally continuity constraints are imposed. 

\notab \textit{Disturbance Observer:} Analogous to the case of LMPC, we can estimate the external disturbances $\mathbf{F}_{ext}$. This is now achieved through an augmented state Extended Kalman Filter (EKF) that includes the external forces. The EKF uses the same model as in control design but further incorporates the heading angle. The external force estimation in turn incorporates modelling errors and supports offset-free tracking.

\notab \textit{Literature Review:}  Beyond this baseline formulation of NMPC for MAVs, the research in the community has investigated further problems. The contribution in~\cite{bicego2020nonlinear} considers general MAV designs and an enhanced actuator model for improved tracking performance. The work in~\cite{kamel2015fast} examines the problem of applying NMPC directly for the inner attitude dynamics of the system. The authors in~\cite{pereira2019nonlinear} present a NMPC approach formulated on the Special Euclidean group SE(3), which has a single optimization layer and offers safe trajectory tracking with obstacle avoidance capacity. The work in~\cite{bicego2019nonlinear} explicitly considers the role of input constraints in NMPC design for multirotor MAVs. Towards agile performance combined with lightweight computational needs, the work in~\cite{neunert2016fast} presented a method for real-time, unconstrained NMPC that combines trajectory optimization and tracking control in a single, unified approach. It uses an iterative optimal control algorithm - namely Sequential Linear Quadratic - in the MPC setting to solve the underlying nonlinear control problem and simultaneously derive the optimal feedforward and feedback terms. The authors demonstrate that the solver can generate trajectories with a duration of multiple seconds within only a few milliseconds. Focusing on the problem of collision-free flight, the contribution in~\cite{garimella2017robust} applies NMPC for the problem of obstacle avoidance for a quadrotor aerial vehicle. Similarly, the work in~\cite{small2019aerial} utilizes NMPC to enable the avoidance of complex obstacles including those with non-convex shape. Considering the specific need of carrying external payloads, the work in~\cite{gonzalez2015non} applies NMPC for slung load oscillation suppression for a quadrotor MAV.

\subsection{Comparison of Linear and Nonlinear MPC}\label{sec:comparison}
As free-flight control is the main control task for a multirotor MAV, in this section we present a comparison of two baseline linear and nonlinear MPC approaches for the position tracking problem of a hexarotor MAV.

%
\begin{figure}[h!]
\centering
    \includegraphics[width=0.95\columnwidth]{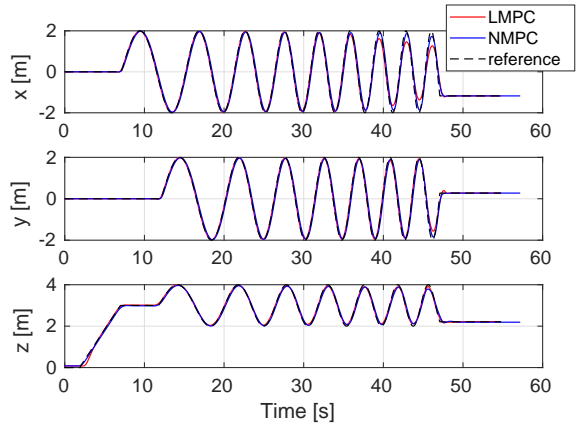}
\caption{Position responses of Linear and Nonlinear MPC with sinusoidal input signal having frequency varied in the range $[0.1,0.33]\textrm{Hz}$.}\label{fig:LMPC_vs_NMPC}
\vspace{-2ex} 
\end{figure}
%

\begin{figure}[h!]
\centering
    \includegraphics[width=0.95\columnwidth]{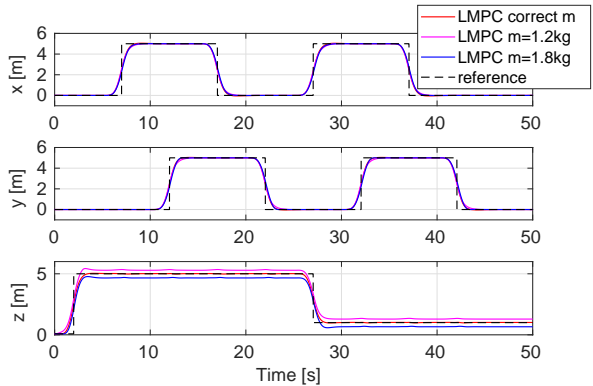}
\caption{Step responses of Linear MPC when the mass of the MAV is set correctly ($m=1.5\textrm{kg}$) and when the mass is incorrect ($m=1.2\textrm{kg}$ and $m=1.8\textrm{kg}$). The disturbance observer is turned off in all cases. RMSE in z-axis are $30.36, 42.3$, and $45.1\textrm{cm}$, respectively.}\label{fig:LMPC_uncertain_mass}
\vspace{-2ex} 
\end{figure}

More specifically, the Linear and Nonlinear MPC controllers' performance are compared using the C++ implementations presented in ~\cite{mpc_rosbookchapter}, with the simulated model being an AscTec Firefly hexacopter based on the RotorS open-source simulator~\cite{furrer2016rotors}. The weight matrices $\mathbf{Q}_x$ and $\mathbf{R}_u$ are chosen the same for both controllers, while the terminal matrix $\mathbf{P}$ is calculated by solving the corresponding discrete algebraic Ricatti equation. From Figure~\ref{fig:LMPC_vs_NMPC}, it is observed that the Nonlinear MPC outperforms the Linear MPC when the trajectory is more aggressive ($t\in [40,48]s$) since the Nonlinear MPC can exploit the nonlinear dynamics of the system when the tilt angles of the drone are large. 
The RMSE errors of the Nonlinear and Linear MPC in this case are $8.6$ and $19.0\textrm{cm}$, respectively. The performance of the linear MPC with parameter uncertainty, in this case the mass parameter, is also verified and the results are illustrated in Figure~\ref{fig:LMPC_uncertain_mass}. It can be seen that even though the responses in $x,y$ axes are not affected much, there is offset in the $z$ axis response when the mass of the system is incorrect which necessitates to incorporate a disturbance observer in practical use. 

%
%

\begin{figure}[h!]
\centering
    \includegraphics[width=0.95\columnwidth]{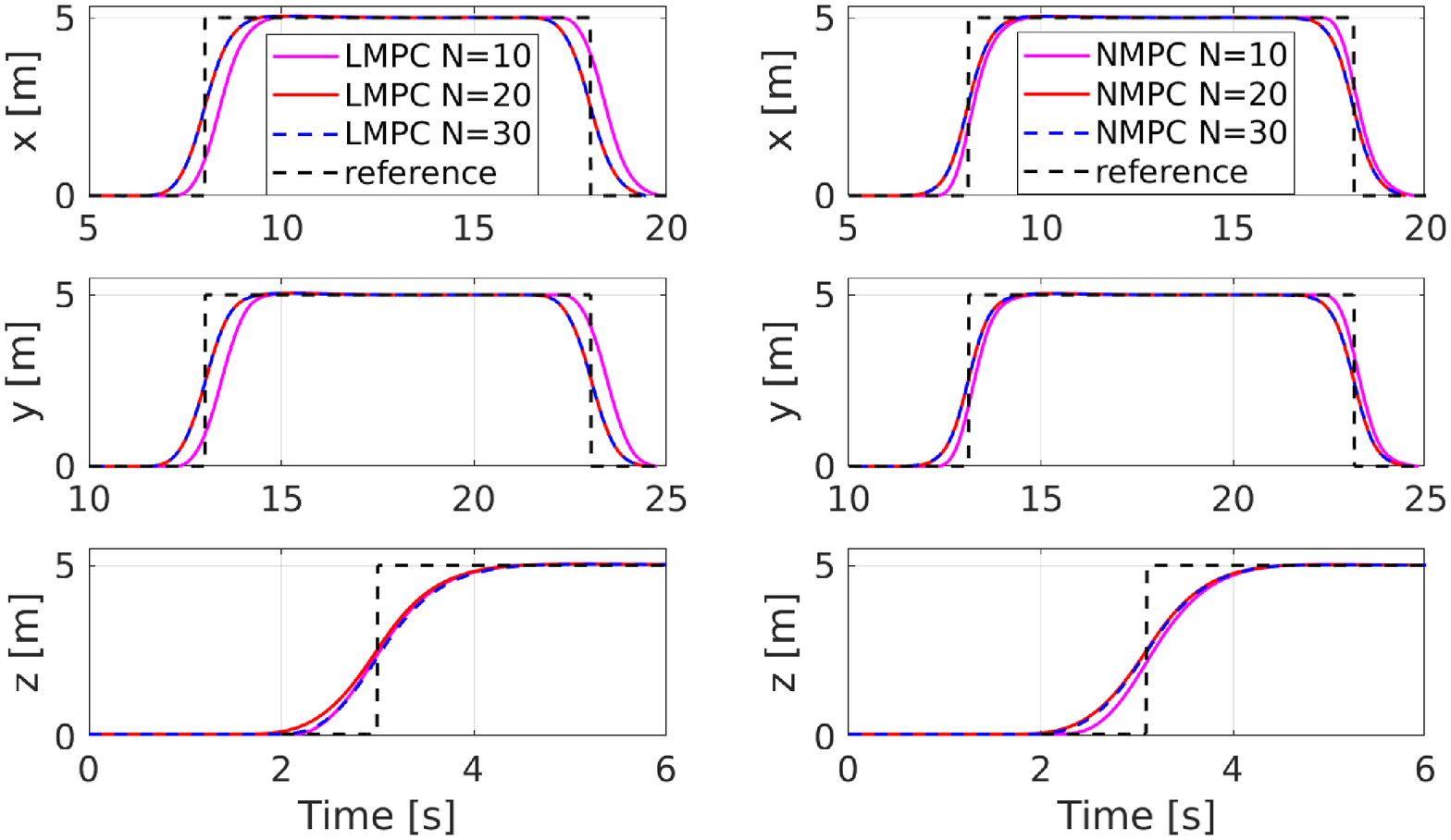}
\caption{Step responses of Linear and Nonlinear MPC with different prediction horizons ($N=10,20,30T_p$ with prediction step $T_p=0.1s$).}\label{fig:LMPC_NMPC_horizon}
\vspace{-2ex} 
\end{figure}

\begin{figure}[h!]
\centering
    \includegraphics[width=0.95\columnwidth]{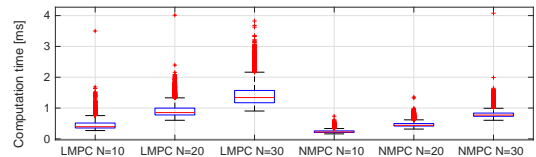}
\caption{Computation time of the control loop of Linear and Nonlinear MPC with different prediction horizons on an i7 8th gen Intel CPU. The reference signal is illustrated in Figure~{\ref{fig:LMPC_uncertain_mass}}.}\label{fig:LMPC_NMPC_computation_time}
\vspace{-2ex} 
\end{figure}

It is known that the number of prediction steps in the MPC problem can greatly affect the feasibility and stability of the closed-loop system. Specifically, increasing the prediction horizon leads to larger region of attraction \cite{borrelli_bemporad_morari_2017}. The responses of the closed loop system with different prediction horizons and input signal described in Figure~\ref{fig:LMPC_uncertain_mass} are illustrated in Figure~\ref{fig:LMPC_NMPC_horizon} and the RMSE values are given in Table~\ref{tab:RMSE}. It can be seen that reasonably increasing the number of prediction steps improves the tracking performance. However, solving the MPC problem with larger prediction horizon requires more computation time as described in the box plot in Figure~{\ref{fig:LMPC_NMPC_computation_time}}. The outlier values denoted by red crosses in Figure~\ref{fig:LMPC_NMPC_computation_time} correspond to the cases when the control inputs are close to the limits, which require the solvers to take more iterations to find the solutions. Interestingly, the nonlinear MPC solver based on~\cite{houska2011acado} has smaller computation time compared to the linear MPC solver based on~\cite{mattingley2012cvxgen}. 

%
%

\begin{table}[h!]
\caption{RMSE values of the $xyz$ responses of Linear and Nonlinear MPC with reference signal given in Figure~\ref{fig:LMPC_uncertain_mass}}
\label{tab:RMSE}
\vspace{-3ex}
\begin{center}
\begin{tabular}{|l|l|l|l|}
\hline
              & $N=10$ & $N=20$ & $N=30$\\ \hline
\textbf{LMPC}~(\textrm{m}) & $1.06$ & $0.78$ & $0.78$\\ \hline
\textbf{NMPC}~(\textrm{m}) & $0.79$ & $0.74$ & $0.74$\\ \hline
\end{tabular}
\end{center}
\end{table} 

%
\vspace{-4ex}
%

\subsection{Fault-Tolerant MPC}\label{sec:ftmpc}
Fault-tolerance is an essential property of every control scheme. As MAVs can undertake critical roles, while their airborne nature makes them potential risk factors, assessing the fault-tolerance of their flight control is particularly important. The work in~\cite{mueller2014stability} has demonstrated the potential to retain full or partial controllability of a quadrotor MAV degrees-of-freedom even subject to the loss of one, two or even three propellers. Naturally, more control re-allocation options arise with MAVs integrating additional actuators (e.g., a hexacopter). In terms of MPC work, the contributions in~\cite{kamel2015fast,tzoumanikas2020nonlinear,wu2019nonlinear} demonstrated - following different designs - the inherent capability of NMPC to retain dynamic stability for a symmetric underactuated hexacopter subject to propeller loss. Furthermore, the contribution in~\cite{aoki2018nonlinear} demonstrates the application of NMPC for a hexarotor with three motor failures. The authors in~\cite{yu2015mpc} investigate the role of partial loss of control effectiveness in the actuators of a quadrotor and apply MPC with terminal constraints to enable the accurate reference tracking despite the considered faults. A fault detection and diagnosis system is designed to assist MPC in its task. It is considered that the importance of integration of MAVs in safety-critical applications or the national airspace will increase the importance of fault-tolerant predictive control design. 

\subsection{Deep Reinforcement Learning}\label{sec:rlmpc}
MPC, which aims to find a solution of the constrained finite-horizon optimization problem, is closely related to Reinforcement Learning (RL), which learns how to make sequential decisions to maximize a numerical reward signal through trial-and-error search~\cite{sutton18RL_introduction}. The interactive nature of RL combined with the approximation ability of neural networks, allow the replacement of each component in the MPC scheme (or part of it) with this powerful representation. The works in~\cite{hoeller2020DMPC,lowrey2018plan} derive the terminal and transition cost functions from the value function which is learned by rolling out the current policy and collecting reward signals. This reward signal can be a binary or sparse reward which opens the opportunity to remove the need for hand-tuning the cost matrices in MPC~\cite{karnchanachari20practicalMPC}. The authors in~\cite{Nagabandi2018NeuralND} use a neural network to learn the dynamic function of the system, while the contribution in~\cite{Fan2020DeepLT} proposes a deep quantile regression framework for learning bounds on distributions of trajectories, demonstrated to generate an obstacle avoidance path for a full-state quadrotor model subject to action noise. The computation cost for solving the MPC problem can be high with long prediction horizon, rendering it impractical to be applied to many real-time control problems and in such cases, deep RL can be used to compress the MPC policy. The work in~\cite{zhang2016MPC_GPS} uses an expert MPC in guided policy search to control a MAV which not only reduces the computation time compared to that of the expert MPC but also removes the need for an explicit state estimation. The authors in~\cite{chen18NN_explicitMPC} propose a constrained neural network architecture to imitate an explicit MPC law and then a policy gradient method - with the advantage function calculated by utilizing the terminal cost function in a MPC problem - is developed. It is noted that the use of neural networks to represent the optimal policy in critical constrained optimization problems necessitates the need for verification methods to validate the performance of the close loop systems. The work in~\cite{Hu2020ReachSDPRA} demonstrates computing the $10$-step forward reachable set of a $6\textrm{D}$ quadrotor model controlled by a neural network using Semidefinite Programming.

\subsection{Load Transportation}\label{sec:tmpc}

Analogous to their manned counterparts, micro aerial vehicles are considered for load transportation tasks~\cite{villa2019survey}. Despite the robustness of MPC and especially of certain design variations of it~\cite{alexis2016robust}, special control design is necessary for high-performance load transportation using one or more multirotor systems. The work in~\cite{jain2015transportation} presents a method for cable-suspended load transportation using a quadorotor vehicle. The authors in~\cite{garimella2015towards} present aerial pick-and-place relying on MPC methods. Considering the benefits of tilt-rotor systems, the works in~\cite{santos2016path,andrade2016model} propose MPC methods for load transportation. As during a slung-load operation, it is not only the aerial robot that can collide with the world but also the load itself, the contribution~\cite{son2018model} explicitly derives safe paths for load transportation operations. Considering the potential of multi-robot synergy in load transportation, a possible MPC design is presented in~\cite{alothman2018using} for two vehicles, while a more general problem formulation is detailed in~\cite{tartaglione2017model}.

\subsection{Physical Interaction}\label{sec:hmpc}

\begin{figure}[h!]
\centering
    \includegraphics[width=0.94\columnwidth]{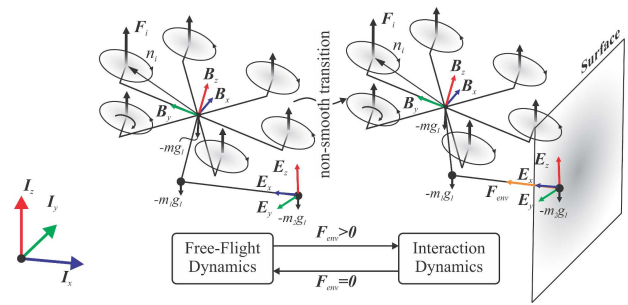}
\caption{Physical interaction with micro aerial vehicles affords hybrid systems formulation. In free-flight the manipulator/end-effector-based induced disturbances should also be accounted, while during physical interaction the forces exerted by the environment have to be considered.}\label{fig:hmpcconcept}
\vspace{-1ex} 
\end{figure}

MPC methods have also found their way in the context of research work relating to aerial robots physically interacting with their environment. The authors in~\cite{DABS_ICRA_14} derive a hybrid systems-based formulation of a quadrotor that either navigates in free-flight or comes in contact with the environment in order to perform inspection tasks. The work first utilizes a linearized model for the position dynamics of the quadrotor in free-flight given the system identification of the closed-loop attitude dynamics. This is combined with a linear model of the system in contact with the environment by accounting for the force applied from the physical surfaces. The applicability of hybrid systems relates to the fact that collision-dynamics are particularly fast and thus allow to handle them as nonsmooth effects instead of stiff differential equations~\cite{jean1999non}. A broader illustration is depicted in Figure~\ref{fig:hmpcconcept}. Utilizing similar principles, the work in~\cite{papachristos2014technical} performs forceful work-tasks using MPC and a tilt-rotor MAV. 

%
%

Investigating a more challenging task, the authors in~\cite{garimella2015towards} proposed a MPC framework for a MAV performing aerial pick-and-place tasks. Examining the problem of aerial manipulation, the authors in~\cite{lunni2017nonlinear} propose a NMPC to follow desired trajectories with the end-effecctor of a multirotor. The work further examines the potential enabled by the augmented kinematics the manipulator offers during free-flight. Considering the explicit task of opening a door, the contribution in~\cite{lee2019model} proposes a model predictive control framework, albeit in simulation, for a quadrotor utilizing an onboard arm to open a hinged door. Extending the potential capacity of a MAV to perform work-tasks in its environment, the work in~\cite{kocer2018constrained} considers the problem of the robot interacting with its environment through an elastic tool.

\section{OPEN-SOURCE MPC PACKAGES FOR MAVs}\label{sec:opensource}
The success of MPC in the problem of trajectory tracking for MAVs is also reflected in the extensive utilization of relevant open-source packages released. The work in~\cite{mpc_rosbookchapter} is associated with an open-source Robot Operating System (ROS) package available at \url{https://github.com/ethz-asl/mav_control_rw} that offers both linear and nonlinear MPC laws. The code in \url{https://github.com/uzh-rpg/rpg_mpc} also provides MPC functionality for multirotors and has extensions to perception-aware functionality~\cite{falanga2018pampc}. Similarly, it is released as a ROS package. The work in~\cite{dentler2016real} is also released as an open-source contribution and provides both multi-robot and single-robot control such as NMPC for quadrotors. It can be found as a ROS package at \url{https://github.com/DentOpt/denmpc}. Last, an implementation for ARM CPUs~\cite{baca2016embedded} can be found at~\url{https://github.com/klaxalk/multirotor-control-board}. Contributing a larger overall software library for control, the work in~\cite{giftthaler2018control} also provides an example for MPC of quadrotors and can be found at \url{https://github.com/ethz-adrl/control-toolbox}. These works are indicative and more are available in the community. Simultaneously, the interested researcher can also directly refer to software packages for general MPC design such as CVXGEN~\cite{mattingley2012cvxgen} available at \url{https://cvxgen.com/docs/index.html}, ACADO~\cite{houska2011acado} available at \url{http://acado.sourceforge.net/doc/html/d4/d26/example_013.html}, YALMIP~\cite{lofberg2004yalmip} available at \url{https://yalmip.github.io/}, the Multi-Parametric Toolbox~\cite{MPT3} available at \url{https://www.mpt3.org/}, do-mpc~\cite{lucia2017rapid} found at \url{https://www.do-mpc.com/en/latest/} and other both open-source and closed packages applicable to a variety of programming languages and processor architectures.

\section{CONCLUSIONS}\label{sec:conclusion}

A survey on the application and design considerations of model predictive control for micro aerial vehicles was presented. The literature in the domain includes both linear and nonlinear controllers for the robot flight dynamics, methods for physical interaction and load transportation, fault-tolerant control schemes and methods combining modern reinforcement learning techniques. As the integration of MAVs in important application domains becomes wider, we anticipate that the study of novel MPC methods - especially considering the uncertainties and risks in the robot navigation process - will tend to be even more important and possibly essential for achieving robust autonomous flight.

\bibliographystyle{IEEEtran}
\bibliography{./BIB/MPCSURVEY_ECC_2021}

\begin{thebibliography}{10}
\providecommand{\url}[1]{#1}
\csname url@samestyle\endcsname
\providecommand{\newblock}{\relax}
\providecommand{\bibinfo}[2]{#2}
\providecommand{\BIBentrySTDinterwordspacing}{\spaceskip=0pt\relax}
\providecommand{\BIBentryALTinterwordstretchfactor}{4}
\providecommand{\BIBentryALTinterwordspacing}{\spaceskip=\fontdimen2\font plus
\BIBentryALTinterwordstretchfactor\fontdimen3\font minus
  \fontdimen4\font\relax}
\providecommand{\BIBforeignlanguage}[2]{{%
\expandafter\ifx\csname l@#1\endcsname\relax
\typeout{** WARNING: IEEEtran.bst: No hyphenation pattern has been}%
\typeout{** loaded for the language `#1'. Using the pattern for}%
\typeout{** the default language instead.}%
\else
\language=\csname l@#1\endcsname
\fi
#2}}
\providecommand{\BIBdecl}{\relax}
\BIBdecl

\bibitem{SIP_AURO_2015}
{A. Bircher, M. Kamel, K. Alexis, M. Burri, P. Oettershagen, S. Omari, T.
  Mantel and R. Siegwart}, ``\BIBforeignlanguage{English}{Three-dimensional
  coverage path planning via viewpoint resampling and tour optimization for
  aerial robots},'' \emph{\BIBforeignlanguage{English}{Autonomous Robots}}, pp.
  1--25, 2015.

\bibitem{grocholsky2006cooperative}
B.~Grocholsky, J.~Keller, V.~Kumar, and G.~Pappas, ``Cooperative air and ground
  surveillance,'' \emph{IEEE Robotics \& Automation Magazine}, vol.~13, no.~3,
  pp. 16--25, 2006.

\bibitem{garimella2015towards}
G.~Garimella and M.~Kobilarov, ``Towards model-predictive control for aerial
  pick-and-place,'' in \emph{2015 IEEE international conference on robotics and
  automation (ICRA)}.\hskip 1em plus 0.5em minus 0.4em\relax IEEE, 2015, pp.
  4692--4697.

\bibitem{agha2014health}
A.-a. Agha-mohammadi, N.~K. Ure, J.~P. How, and J.~Vian, ``Health aware
  stochastic planning for persistent package delivery missions using
  quadrotors,'' in \emph{2014 IEEE/RSJ International Conference on Intelligent
  Robots and Systems}.\hskip 1em plus 0.5em minus 0.4em\relax IEEE, 2014, pp.
  3389--3396.

\bibitem{camacho2013model}
E.~F. Camacho and C.~B. Alba, \emph{Model predictive control}.\hskip 1em plus
  0.5em minus 0.4em\relax Springer Science \& Business Media, 2013.

\bibitem{allgower2012nonlinear}
F.~Allg{\"o}wer and A.~Zheng, \emph{Nonlinear model predictive control}.\hskip
  1em plus 0.5em minus 0.4em\relax Birkh{\"a}user, 2012, vol.~26.

\bibitem{mayne2000constrained}
D.~Q. Mayne, J.~B. Rawlings, C.~V. Rao, and P.~O. Scokaert, ``Constrained model
  predictive control: Stability and optimality,'' \emph{Automatica}, vol.~36,
  no.~6, pp. 789--814, 2000.

\bibitem{grune2017nonlinear}
L.~Gr{\"u}ne and J.~Pannek, ``Nonlinear model predictive control,'' in
  \emph{Nonlinear Model Predictive Control}.\hskip 1em plus 0.5em minus
  0.4em\relax Springer, 2017, pp. 45--69.

\bibitem{garcia1989model}
C.~E. Garcia, D.~M. Prett, and M.~Morari, ``Model predictive control: theory
  and practice—a survey,'' \emph{Automatica}, vol.~25, no.~3, 1989.

\bibitem{rawlings2000tutorial}
J.~B. Rawlings, ``Tutorial overview of model predictive control,'' \emph{IEEE
  control systems magazine}, vol.~20, no.~3, pp. 38--52, 2000.

\bibitem{bemporad1999robust}
A.~Bemporad and M.~Morari, ``Robust model predictive control: A survey,'' in
  \emph{Robustness in identification and control}.\hskip 1em plus 0.5em minus
  0.4em\relax Springer, 1999.

\bibitem{falanga2018pampc}
D.~Falanga, P.~Foehn, P.~Lu, and D.~Scaramuzza, ``Pampc: Perception-aware model
  predictive control for quadrotors,'' in \emph{2018 IEEE/RSJ International
  Conference on Intelligent Robots and Systems (IROS)}.\hskip 1em plus 0.5em
  minus 0.4em\relax IEEE, 2018, pp. 1--8.

\bibitem{hoffmann2007quadrotor}
G.~Hoffmann, H.~Huang, S.~Waslander, and C.~Tomlin, ``Quadrotor helicopter
  flight dynamics and control: Theory and experiment,'' in \emph{AIAA guidance,
  navigation and control conference and exhibit}, 2007.

\bibitem{mahony2012multirotor}
R.~Mahony, V.~Kumar, and P.~Corke, ``Multirotor aerial vehicles: Modeling,
  estimation, and control of quadrotor,'' \emph{IEEE Robotics and Automation
  magazine}, vol.~19, no.~3, pp. 20--32, 2012.

\bibitem{omari2013nonlinear}
S.~Omari, M.-D. Hua, G.~Ducard, and T.~Hamel, ``Nonlinear control of vtol uavs
  incorporating flapping dynamics,'' in \emph{2013 IEEE/RSJ International
  Conference on Intelligent Robots and Systems}, 2013.

\bibitem{kamel2015fast}
M.~Kamel, K.~Alexis, M.~Achtelik, and R.~Siegwart, ``Fast nonlinear model
  predictive control for multicopter attitude tracking on so (3),'' in
  \emph{2015 IEEE Conference on Control Applications (CCA)}.\hskip 1em plus
  0.5em minus 0.4em\relax IEEE, 2015, pp. 1160--1166.

\bibitem{lee2010geometric}
T.~Lee, M.~Leok, and N.~H. McClamroch, ``Geometric tracking control of a
  quadrotor uav on se (3),'' in \emph{49th IEEE conference on decision and
  control (CDC)}.\hskip 1em plus 0.5em minus 0.4em\relax IEEE, 2010, pp.
  5420--5425.

\bibitem{kamel2017linear}
M.~Kamel, M.~Burri, and R.~Siegwart, ``Linear vs nonlinear mpc for trajectory
  tracking applied to rotary wing micro aerial vehicles,''
  \emph{IFAC-PapersOnLine}, vol.~50, no.~1, pp. 3463--3469, 2017.

\bibitem{alexis2010design}
K.~Alexis, G.~Nikolakopoulos, and A.~Tzes, ``Design and experimental
  verification of a constrained finite time optimal control scheme for the
  attitude control of a quadrotor helicopter subject to wind gusts,'' in
  \emph{2010 IEEE International Conference on Robotics and Automation}.\hskip
  1em plus 0.5em minus 0.4em\relax IEEE, 2010, pp. 1636--1641.

\bibitem{MPT3}
M.~Herceg, M.~Kvasnica, C.~Jones, and M.~Morari, ``{Multi-Parametric Toolbox
  3.0},'' in \emph{Proc.~of the European Control Conference}, Z\"urich,
  Switzerland, July 17--19 2013, \url{http://control.ee.ethz.ch/~mpt}.

\bibitem{raffo2010integral}
G.~V. Raffo, M.~G. Ortega, and F.~R. Rubio, ``An integral predictive/nonlinear
  h-infinity control structure for a quadrotor helicopter,'' \emph{Automatica},
  vol.~46, no.~1, pp. 29--39, 2010.

\bibitem{alexis2011switching}
K.~Alexis, G.~Nikolakopoulos, and A.~Tzes, ``Switching model predictive
  attitude control for a quadrotor helicopter subject to atmospheric
  disturbances,'' \emph{Control Engineering Practice}, vol.~19, no.~10, pp.
  1195--1207, 2011.

\bibitem{alexis2012model}
{K. Alexis, G. Nikolakopoulos, and A. Tzes}, ``Model predictive quadrotor
  control: attitude, altitude and position experimental studies,'' \emph{IET
  Control Theory \& Applications}, vol.~6, no.~12, pp. 1812--1827, 2012.

\bibitem{alexis2016robust}
K.~Alexis, C.~Papachristos, R.~Siegwart, and A.~Tzes, ``Robust model predictive
  flight control of unmanned rotorcrafts,'' \emph{Journal of Intelligent \&
  Robotic Systems}, vol.~81, no. 3-4, pp. 443--469, 2016.

\bibitem{greeff2018flatness}
M.~Greeff and A.~P. Schoellig, ``Flatness-based model predictive control for
  quadrotor trajectory tracking,'' in \emph{2018 IEEE/RSJ International
  Conference on Intelligent Robots and Systems (IROS)}.\hskip 1em plus 0.5em
  minus 0.4em\relax IEEE, 2018.

\bibitem{schurmann2018reachset}
B.~Sch{\"u}rmann, N.~Kochdumper, and M.~Althoff, ``Reachset model predictive
  control for disturbed nonlinear systems,'' in \emph{2018 IEEE Conference on
  Decision and Control (CDC)}.\hskip 1em plus 0.5em minus 0.4em\relax IEEE,
  2018.

\bibitem{gillula2011applications}
J.~H. Gillula, G.~M. Hoffmann, H.~Huang, M.~P. Vitus, and C.~J. Tomlin,
  ``Applications of hybrid reachability analysis to robotic aerial vehicles,''
  \emph{The International Journal of Robotics Research}, 2011.

\bibitem{aswani2012extensions}
A.~Aswani, P.~Bouffard, and C.~Tomlin, ``Extensions of learning-based model
  predictive control for real-time application to a quadrotor helicopter,'' in
  \emph{2012 American Control Conference (ACC)}.\hskip 1em plus 0.5em minus
  0.4em\relax IEEE, 2012, pp. 4661--4666.

\bibitem{bicego2020nonlinear}
D.~Bicego, J.~Mazzetto, R.~Carli, M.~Farina, and A.~Franchi, ``Nonlinear model
  predictive control with enhanced actuator model for multi-rotor aerial
  vehicles with generic designs,'' \emph{Journal of Intelligent \& Robotic
  Systems}, pp. 1--35, 2020.

\bibitem{pereira2019nonlinear}
J.~C. Pereira, V.~J. Leite, and G.~V. Raffo, ``Nonlinear model predictive
  control on se (3) for quadrotor trajectory tracking and obstacle avoidance,''
  in \emph{2019 19th International Conference on Advanced Robotics
  (ICAR)}.\hskip 1em plus 0.5em minus 0.4em\relax IEEE, 2019, pp. 155--160.

\bibitem{bicego2019nonlinear}
D.~Bicego, J.~Mazzetto, R.~Carli, M.~Farina, and A.~Franchi, ``Nonlinear model
  predictive control with actuator constraints for multi-rotor aerial
  vehicles,'' \emph{arXiv preprint arXiv:1911.08183}, 2019.

\bibitem{neunert2016fast}
M.~Neunert, C.~De~Crousaz, F.~Furrer, M.~Kamel, F.~Farshidian, R.~Siegwart, and
  J.~Buchli, ``Fast nonlinear model predictive control for unified trajectory
  optimization and tracking,'' in \emph{2016 IEEE international conference on
  robotics and automation (ICRA)}.

\bibitem{garimella2017robust}
G.~Garimella, M.~Sheckells, and M.~Kobilarov, ``Robust obstacle avoidance for
  aerial platforms using adaptive model predictive control,'' in \emph{2017
  IEEE International Conference on Robotics and Automation (ICRA)}.\hskip 1em
  plus 0.5em minus 0.4em\relax IEEE, 2017, pp. 5876--5882.

\bibitem{small2019aerial}
E.~Small, P.~Sopasakis, E.~Fresk, P.~Patrinos, and G.~Nikolakopoulos, ``Aerial
  navigation in obstructed environments with embedded nonlinear model
  predictive control,'' in \emph{2019 18th European Control Conference
  (ECC)}.\hskip 1em plus 0.5em minus 0.4em\relax IEEE, 2019, pp. 3556--3563.

\bibitem{gonzalez2015non}
F.~Gonzalez, A.~Heckmann, S.~Notter, M.~Z{\"u}rn, J.~Trachte, and A.~McFadyen,
  ``Non-linear model predictive control for uavs with slung/swung load,'' 2015.

\bibitem{mpc_rosbookchapter}
M.~Kamel, T.~Stastny, K.~Alexis, and R.~Siegwart, \emph{Model Predictive
  Control for Trajectory Tracking of Unmanned Aerial Vehicles Using Robot
  Operating System}.\hskip 1em plus 0.5em minus 0.4em\relax Springer
  International Publishing, 2017.

\bibitem{furrer2016rotors}
F.~Furrer, M.~Burri, M.~Achtelik, and R.~Siegwart, ``Rotors—a modular gazebo
  mav simulator framework,'' in \emph{Robot Operating System (ROS)}.\hskip 1em
  plus 0.5em minus 0.4em\relax Springer, 2016, pp. 595--625.

\bibitem{borrelli_bemporad_morari_2017}
F.~Borrelli, A.~Bemporad, and M.~Morari, \emph{Predictive Control for Linear
  and Hybrid Systems}.\hskip 1em plus 0.5em minus 0.4em\relax Cambridge
  University Press, 2017.

\bibitem{houska2011acado}
B.~Houska, H.~J. Ferreau, and M.~Diehl, ``Acado toolkit—an open-source
  framework for automatic control and dynamic optimization,'' \emph{Optimal
  Control Applications and Methods}, vol.~32, no.~3, 2011.

\bibitem{mattingley2012cvxgen}
J.~Mattingley and S.~Boyd, ``Cvxgen: A code generator for embedded convex
  optimization,'' \emph{Optimization and Engineering}, vol.~13, no.~1.

\bibitem{mueller2014stability}
M.~W. Mueller and R.~D'Andrea, ``Stability and control of a quadrocopter
  despite the complete loss of one, two, or three propellers,'' in \emph{2014
  IEEE international conference on robotics and automation (ICRA)}.\hskip 1em
  plus 0.5em minus 0.4em\relax IEEE, 2014, pp. 45--52.

\bibitem{tzoumanikas2020nonlinear}
D.~Tzoumanikas, Q.~Yan, and S.~Leutenegger, ``Nonlinear mpc with motor failure
  identification and recovery for safe and aggressive multicopter flight,''
  \emph{arXiv preprint arXiv:2002.06598}, 2020.

\bibitem{wu2019nonlinear}
Y.~Wu, K.~Hu, X.-M. Sun, and Y.~Ma, ``Nonlinear control of quadrotor for fault
  tolerance: A total failure of one actuator,'' \emph{IEEE Transactions on
  Systems, Man, and Cybernetics: Systems}, 2019.

\bibitem{aoki2018nonlinear}
Y.~Aoki, Y.~Asano, A.~Honda, N.~Motooka, and T.~Ohtsuka, ``Nonlinear model
  predictive control of position and attitude in a hexacopter with three failed
  rotors,'' \emph{IFAC-PapersOnLine}, vol.~51, no.~20, 2018.

\bibitem{yu2015mpc}
B.~Yu, Y.~Zhang, and Y.~Qu, ``Mpc-based ftc with fdd against actuator faults of
  uavs,'' in \emph{2015 15th International Conference on Control, Automation
  and Systems (ICCAS)}.\hskip 1em plus 0.5em minus 0.4em\relax IEEE, 2015, pp.
  225--230.

\bibitem{sutton18RL_introduction}
R.~S. Sutton and A.~G. Barto, \emph{Reinforcement Learning: An
  Introduction}.\hskip 1em plus 0.5em minus 0.4em\relax Cambridge, MA, USA: A
  Bradford Book, 2018.

\bibitem{hoeller2020DMPC}
D.~Hoeller, F.~Farshidian, and M.~Hutter, ``Deep value model predictive
  control,'' 2020, pp. 990--1004.

\bibitem{lowrey2018plan}
K.~Lowrey, A.~Rajeswaran, S.~Kakade, E.~Todorov, and I.~Mordatch, ``Plan
  online, learn offline: Efficient learning and exploration via model-based
  control,'' in \emph{International Conference on Learning Representations},
  2019.

\bibitem{karnchanachari20practicalMPC}
N.~Karnchanachari, M.~de~la Iglesia~Valls, D.~Hoeller, and M.~Hutter,
  ``Practical reinforcement learning for mpc: Learning from sparse objectives
  in under an hour on a real robot,'' ser. Proceedings of Machine Learning
  Research, vol. 120.\hskip 1em plus 0.5em minus 0.4em\relax PMLR, 2020, pp.
  211--224.

\bibitem{Nagabandi2018NeuralND}
A.~Nagabandi, G.~Kahn, R.~S. Fearing, and S.~Levine, ``Neural network dynamics
  for model-based deep reinforcement learning with model-free fine-tuning,''
  \emph{2018 IEEE International Conference on Robotics and Automation (ICRA)},
  pp. 7559--7566, 2018.

\bibitem{Fan2020DeepLT}
D.~D. Fan, A.~akbar Agha-mohammadi, and E.~A. Theodorou, ``Deep learning tubes
  for tube mpc,'' \emph{ArXiv}, vol. abs/2002.01587, 2020.

\bibitem{zhang2016MPC_GPS}
T.~{Zhang}, G.~{Kahn}, S.~{Levine}, and P.~{Abbeel}, ``Learning deep control
  policies for autonomous aerial vehicles with mpc-guided policy search,'' in
  \emph{2016 IEEE International Conference on Robotics and Automation (ICRA)},
  2016, pp. 528--535.

\bibitem{chen18NN_explicitMPC}
S.~{Chen}, K.~{Saulnier}, N.~{Atanasov}, D.~D. {Lee}, V.~{Kumar}, G.~J.
  {Pappas}, and M.~{Morari}, ``Approximating explicit model predictive control
  using constrained neural networks,'' in \emph{2018 Annual American Control
  Conference (ACC)}, 2018, pp. 1520--1527.

\bibitem{Hu2020ReachSDPRA}
H.~Hu, M.~Fazlyab, M.~Morari, and G.~J. Pappas, ``Reach-sdp: Reachability
  analysis of closed-loop systems with neural network controllers via
  semidefinite programming,'' \emph{ArXiv}, vol. abs/2004.07876, 2020.

\bibitem{villa2019survey}
D.~K. Villa, A.~S. Brandao, and M.~Sarcinelli-Filho, ``A survey on load
  transportation using multirotor uavs,'' \emph{Journal of Intelligent \&
  Robotic Systems}, pp. 1--30, 2019.

\bibitem{jain2015transportation}
R.~P.~K. Jain, ``Transportation of cable suspended load using unmanned aerial
  vehicles: A real-time model predictive control approach,'' 2015.

\bibitem{santos2016path}
M.~A. Santos and G.~V. Raffo, ``Path tracking model predictive control of a
  tilt-rotor uav carrying a suspended load,'' in \emph{2016 IEEE 19th
  international conference on intelligent transportation systems}, 2016.

\bibitem{andrade2016model}
R.~Andrade, G.~V. Raffo, and J.~E. Normey-Rico, ``Model predictive control of a
  tilt-rotor uav for load transportation,'' in \emph{2016 European Control
  Conference (ECC)}.\hskip 1em plus 0.5em minus 0.4em\relax IEEE, 2016, pp.
  2165--2170.

\bibitem{son2018model}
C.~Y. Son, H.~Seo, T.~Kim, and H.~J. Kim, ``Model predictive control of a
  multi-rotor with a suspended load for avoiding obstacles,'' in \emph{2018
  IEEE International Conference on Robotics and Automation (ICRA)}.

\bibitem{alothman2018using}
Y.~Alothman and D.~Gu, ``Using constrained model predictive control to control
  two quadrotors transporting a cable-suspended payload,'' in \emph{2018 13th
  World Congress on Intelligent Control and Automation (WCICA)}.\hskip 1em plus
  0.5em minus 0.4em\relax IEEE, 2018, pp. 228--233.

\bibitem{tartaglione2017model}
G.~Tartaglione, E.~D’Amato, M.~Ariola, P.~S. Rossi, and T.~A. Johansen,
  ``Model predictive control for a multi-body slung-load system,''
  \emph{Robotics and Autonomous Systems}, vol.~92, pp. 1--11, 2017.

\bibitem{DABS_ICRA_14}
G.~Darivianakis, K.~Alexis, M.~Burri, and R.~Siegwart, ``Hybrid predictive
  control for aerial robotic physical interaction towards inspection
  operations,'' in \emph{Robotics and Automation (ICRA), 2014 IEEE
  International Conference on}, May 2014, pp. 53--58.

\bibitem{jean1999non}
M.~Jean, ``The non-smooth contact dynamics method,'' 1999.

\bibitem{papachristos2014technical}
C.~Papachristos, K.~Alexis, and A.~Tzes, ``Technical activities execution with
  a tiltrotor uas employing explicit model predictive control,'' \emph{IFAC
  Proceedings Volumes}, vol.~47, no.~3, pp. 11\,036--11\,042, 2014.

\bibitem{lunni2017nonlinear}
D.~Lunni, A.~Santamaria-Navarro, R.~Rossi, P.~Rocco, L.~Bascetta, and
  J.~Andrade-Cetto, ``Nonlinear model predictive control for aerial
  manipulation,'' in \emph{2017 International Conference on Unmanned Aircraft
  Systems (ICUAS)}.\hskip 1em plus 0.5em minus 0.4em\relax IEEE, 2017, pp.
  87--93.

\bibitem{lee2019model}
D.~Lee, D.~Jang, H.~Seo, and H.~J. Kim, ``Model predictive control for an
  aerial manipulator opening a hinged door,'' in \emph{2019 19th International
  Conference on Control, Automation and Systems (ICCAS)}.\hskip 1em plus 0.5em
  minus 0.4em\relax IEEE, 2019, pp. 986--991.

\bibitem{kocer2018constrained}
B.~B. Kocer, T.~Tjahjowidodo, and G.~S.~G. Lee, ``Constrained estimation-based
  nonlinear model predictive control for uav-elastic tool interaction,'' in
  \emph{2018 IEEE/ASME International Conference on Advanced Intelligent
  Mechatronics (AIM)}.\hskip 1em plus 0.5em minus 0.4em\relax IEEE, 2018, pp.
  466--471.

\bibitem{dentler2016real}
J.~Dentler, S.~Kannan, M.~A.~O. Mendez, and H.~Voos, ``A real-time model
  predictive position control with collision avoidance for commercial low-cost
  quadrotors,'' in \emph{2016 IEEE conference on control applications
  (CCA)}.\hskip 1em plus 0.5em minus 0.4em\relax IEEE, 2016, pp. 519--525.

\bibitem{baca2016embedded}
T.~Baca, G.~Loianno, and M.~Saska, ``Embedded model predictive control of
  unmanned micro aerial vehicles,'' in \emph{2016 21st international conference
  on methods and models in automation and robotics (MMAR)}.\hskip 1em plus
  0.5em minus 0.4em\relax IEEE, 2016, pp. 992--997.

\bibitem{giftthaler2018control}
M.~Giftthaler, M.~Neunert, M.~St{\"a}uble, and J.~Buchli, ``The control
  toolbox—an open-source c++ library for robotics, optimal and model
  predictive control,'' in \emph{2018 IEEE International Conference on
  Simulation, Modeling, and Programming for Autonomous Robots (SIMPAR)}.\hskip
  1em plus 0.5em minus 0.4em\relax IEEE, 2018, pp. 123--129.

\bibitem{lofberg2004yalmip}
J.~Lofberg, ``Yalmip: A toolbox for modeling and optimization in matlab,'' in
  \emph{2004 IEEE international conference on robotics and automation (IEEE
  Cat. No. 04CH37508)}.\hskip 1em plus 0.5em minus 0.4em\relax IEEE, 2004, pp.
  284--289.

\bibitem{lucia2017rapid}
S.~Lucia, A.~T{\u{a}}tulea-Codrean, C.~Schoppmeyer, and S.~Engell, ``Rapid
  development of modular and sustainable nonlinear model predictive control
  solutions,'' \emph{Control Engineering Practice}, vol.~60, 2017.

\end{thebibliography}

\end{document}